\documentclass{article}

\usepackage[preprint]{neurips_2025}
\usepackage{booktabs}
\usepackage{tabularx}
\usepackage{ragged2e}  
\usepackage{amsmath}
\usepackage{enumitem}
\usepackage{amssymb} 
\usepackage{amsthm}
\usepackage{graphicx}
\usepackage{microtype}
\usepackage{adjustbox}
\usepackage{multicol}
\usepackage{multirow}
\usepackage{colortbl} 
\usepackage{pifont}    
\usepackage{tcolorbox}
\usepackage{wrapfig}
\usepackage[table]{xcolor}
\usepackage{adjustbox}

\usepackage[utf8]{inputenc} 
\usepackage[T1]{fontenc}    
\usepackage{hyperref}       
\usepackage{url}            
\usepackage{amsfonts}       
\usepackage{nicefrac}       
\usepackage{xcolor}         

\title{WISTERIA: Learning Clinical Representations from Noisy Supervision via Multi-View Consistency in Electronic Health Records}

\author{
Ruan Dong$^1$, Yuanyun Zhang$^1$, Shi Li$^2$\\
$^1$ University of Science and Technology of China\\
$^2$ Columbia University
}

\newcolumntype{Y}{>{\RaggedRight\arraybackslash}X}

\begin{document}

\maketitle

\begin{abstract}
Representation learning in electronic health records (EHR) has largely followed paradigms inherited from natural language processing, relying on sequence modeling and reconstruction-based objectives that treat clinical labels as ground truth. However, real-world clinical supervision is inherently weak, arising from heterogeneous, noisy, and institution-specific labeling processes such as billing codes, heuristic phenotypes, and incomplete annotations. In this work, we propose WISTERIA, a weakly-supervised representation learning framework that models labels as stochastic observations of an underlying latent clinical state. Instead of optimizing against a single supervision signal, WISTERIA constructs multiple weak supervision operators and learns representations by enforcing consistency across their induced label distributions. This multi-view formulation induces an implicit denoising mechanism, allowing the model to recover clinically meaningful structure by reconciling disagreement between noisy labelers. We further incorporate ontology-aware regularization in the label space to impose semantic structure over supervision signals. Empirically, WISTERIA improves predictive performance across standard EHR benchmarks, demonstrates strong robustness to label noise, and exhibits superior cross-institutional generalization compared to sequence-based pretraining objectives. These results suggest that explicitly modeling the supervision process—rather than treating labels as fixed targets—provides a more appropriate inductive bias for learning robust and clinically meaningful representations from EHR data.
\end{abstract}

\section{Introduction}

Recent progress in healthcare foundation models has largely adhered to a now-familiar paradigm: increasing model scale, pretraining on large heterogeneous corpora, and transferring learned representations to downstream clinical tasks \cite{vaid2023foundational, burkhart2025foundation, thieme2023foundation, he2024foundation}. 
This paradigm has been successfully instantiated across modalities, including structured electronic health records (EHR), imaging, and physiological signals, with a growing emphasis on scale, modality diversity, and pretraining objectives \cite{burger2025foundation, guo2025foundation, liang2024foundation, awais2025foundation, thakur2024foundation}. 
In EHR modeling, patient histories are typically represented as sequences of discrete tokens—diagnoses, procedures, medications, and lab events—processed by architectures adapted from language modeling \cite{he2022masked, devlin2019bert}. 
These models inherit architectural principles from deep learning in vision and NLP, including residual connections, hierarchical feature extraction, and attention-based sequence modeling \cite{he2017multi, he2015deepresiduallearningimage, he2019bag}. 
While effective, this formulation implicitly assumes that clinical information is fully captured by the order and identity of observed events.

This assumption becomes particularly limiting under weak supervision, which is the dominant regime in real-world clinical data. Clinical labels are rarely direct measurements of physiological state; instead, they are noisy proxies derived from billing codes, heuristic phenotypes, or institution-specific annotation pipelines. As a result, supervision is sparse, biased, and heterogeneous, reflecting the process of documentation rather than the underlying clinical reality. Sequence-based pretraining objectives absorb these artifacts, often entangling clinically meaningful structure with noise induced by coding practices.

In this work, we introduce \textbf{WISTERIA} (\textbf{W}eakly-supervised \textbf{I}nference of \textbf{S}tructured la\textbf{T}ent \textbf{E}mbeddings for cl\textbf{R}inical represent\textbf{A}tions), a new paradigm for EHR representation learning that explicitly models supervision as a stochastic, multi-view process. Rather than treating labels as fixed targets, WISTERIA treats them as noisy observations generated from an unobserved latent clinical state.

Formally, let \(z \in \mathcal{Z}\) denote a latent representation of patient physiology, and let \(x \in \mathcal{X}\) denote the observed EHR. We posit that clinical labels \(y \in \mathcal{Y}\) are generated through a noisy channel:
\[
y \sim p(y \mid z; \psi),
\]
where \(\psi\) parameterizes institution-specific labeling functions. Under this view, observed labels are not ground truth, but stochastic realizations of a broader supervision process.

To operationalize this idea, we construct a set of \emph{weak supervision operators} \(\{\mathcal{W}_k\}_{k=1}^K\), each mapping a patient record \(x\) to a pseudo-label distribution \(\tilde{y}_k = \mathcal{W}_k(x)\). These operators capture diverse sources of supervision, including heuristic phenotyping rules, ontology-based label propagation, co-occurrence statistics, and cross-modal agreement signals. Each operator provides a noisy and partial view of the latent state \(z\), and no single operator is assumed to be correct.

We learn representations by enforcing consistency across these supervision views. Let \(h_\theta(x)\) denote the learned embedding, and let \(g_{\theta,k}\) be a prediction head associated with operator \(\mathcal{W}_k\). The training objective is:
\[
\mathcal{L}_{\mathrm{WISTERIA}} =
\sum_{k=1}^K \mathbb{E}_{x}
\left[
\ell\big(g_{\theta,k}(h_\theta(x)),\, \tilde{y}_k\big)
\right]
+ \lambda \sum_{k \neq k'} 
\mathbb{E}_{x}
\left[
D\big(g_{\theta,k}(h_\theta(x)),\, g_{\theta,k'}(h_\theta(x))\big)
\right],
\]
where \(\ell\) is a divergence (e.g., cross-entropy) and \(D\) enforces agreement between supervision views. This objective encourages the representation to capture structure that is stable across heterogeneous, noisy supervision signals.

This formulation induces an implicit denoising mechanism. Each operator \(\mathcal{W}_k\) can be interpreted as a noisy estimator of \(p(y \mid z)\); enforcing agreement across operators recovers a shared latent signal under a multi-view consistency principle. Unlike conventional supervised learning, which treats labels as fixed, WISTERIA models the generative process of supervision itself, allowing the representation to disentangle signal from annotation noise.

To incorporate clinical structure, we introduce a \emph{label-space geometry regularizer}. Let \(\mathcal{Y}\) be endowed with a graph \(G = (V, E)\) derived from clinical ontologies. Predictions are regularized to be smooth with respect to the graph Laplacian \(L\):
\[
\mathcal{L}_{\mathrm{graph}} =
\sum_{k=1}^K 
\mathbb{E}_{x}
\left[
g_{\theta,k}(h_\theta(x))^\top L \, g_{\theta,k}(h_\theta(x))
\right].
\]
This encourages semantically related clinical concepts to have consistent representations, enabling generalization across coding systems and improving robustness in long-tail settings.

A key conceptual shift in WISTERIA is that invariance is enforced in the supervision space rather than the input space. Instead of relying on data augmentations over EHR sequences, we construct multiple noisy supervisory signals and require the representation to reconcile them. This is particularly well-suited to clinical data, where input perturbations are difficult to define, but supervision signals are abundant and diverse.

Moreover, WISTERIA introduces a new scaling axis: the number and diversity of weak supervision operators. Increasing \(K\) provides additional constraints on the latent representation, improving identifiability and robustness without requiring additional labeled data. This complements traditional scaling strategies based on dataset size and model capacity \cite{he2015deepresiduallearningimage, he2019bag, he2017multi}, but operates by scaling supervision rather than data.

In summary, WISTERIA makes the following contributions. It reframes EHR representation learning as inference over latent clinical states under weak, noisy supervision. It introduces a multi-view consistency objective that aggregates heterogeneous supervisory signals. It incorporates ontology-aware regularization directly in the label space, enabling structured reasoning over clinical concepts. Finally, it demonstrates that modeling supervision as a stochastic process yields representations that are more robust to noise, bias, and sparsity, suggesting a new direction for healthcare foundation models beyond sequence-centric pretraining.

\section{Related Works}

Self-supervised learning (SSL) has become the dominant paradigm for representation learning in healthcare, enabling the emergence of foundation models across structured EHRs, clinical text, imaging, and physiological signals \cite{lowelatent, zhang2025collection, lin2025case, huiliang2025clio, ran2025structured, zhang2025chronoformer, lee2024can, chou2025serialized, zhou2026uncertainty, zhang2026discriminative, zhang2026learning}. 
These approaches follow a now-standard recipe: large-scale pretraining on unlabeled data produces representations that transfer effectively to downstream clinical tasks. 
This trajectory mirrors developments in natural language processing and computer vision, where masked modeling and generative objectives \cite{devlin2019bert, he2022masked, lee2025himae}, combined with deep architectures and scaling laws \cite{he2017multi, he2015deepresiduallearningimage, he2019bag}, have yielded broadly useful embeddings. 
Within EHR modeling, however, these methods typically operate over tokenized sequences of clinical events, implicitly assuming that the observed coding of patient data provides a stable and sufficient basis for representation learning.

A substantial body of work has focused on modeling the \emph{structure} of EHR data, particularly its temporal and hierarchical organization. 
Sequence-based architectures, including transformer variants such as Chronoformer, explicitly encode temporal dependencies and long-range interactions within patient histories \cite{zhang2025chronoformer, zhang2025collection, ran2025structured}. 
Related efforts extend these ideas to high-frequency physiological signals, drawing on classical signal processing priors \cite{oppenheim1999discrete, daubechies1992ten} and large-scale representation learning for biosignals \cite{abbaspourazad2023large, abbaspourazad2024wearable, yang2023biot, lee2025foundation, lee2025himae}. 
While these approaches improve predictive performance by capturing temporal dynamics, they remain fundamentally tied to the observed data modality, treating sequences or signals as the primary object of modeling. 
In contrast, our formulation does not assume that the structure of the input alone determines the representation; instead, it treats both inputs and labels as partial observations of a latent clinical state, and focuses on reconciling heterogeneous supervisory signals.

Another major direction centers on \emph{generative and predictive modeling} of EHR data. 
Autoregressive models and masked reconstruction objectives aim to approximate the distribution of observed clinical records, learning representations by predicting missing or future events \cite{brown2020language}. 
These methods have been widely applied to tasks such as diagnosis prediction, risk stratification, and synthetic data generation \cite{lee2025clinical,steinberg2021language, rasmy2021med, wornow2023shaky, fallahpour2024ehrmamba, lee2024emergency, ono2024text, steinberg2024motor, lee2025using}. 
However, such approaches implicitly equate statistical fidelity with semantic fidelity: accurately reconstructing observed codes is assumed to yield clinically meaningful representations. 
This assumption is problematic in the presence of noisy and weak supervision, where observed labels and codes reflect institutional processes rather than ground-truth physiology. 
Our approach departs from this paradigm by explicitly modeling supervision as a stochastic process and learning representations that are consistent across multiple noisy labeling mechanisms, rather than optimizing for reconstruction accuracy.

Closely related are methods based on \emph{contrastive and alignment-based learning}, which enforce invariance across multiple views of the same data \cite{tian_2019_contrastic_distillation, bertram2024contrastivelearningpreferencescontextual}. 
These techniques have been extended to clinical settings through multimodal alignment, shared embedding spaces, and cross-attention architectures \cite{chen2021crossvit, hou2019cross, huang2019ccnet, wornow2024context, odgaard2024core, shmatko2025learning}. 
The broader paradigm has been highly successful in vision and multimodal learning \cite{radford2021learning, caron2021emerging, zhou2021ibot, saharia2022photorealistic, rombach2021highresolution, oquab2023dinov2, Ranftl2022, kirillov2023segment, liu2024visual, lin2025case}, where invariance to augmentations yields semantically meaningful representations. 
Our work is conceptually aligned with this perspective but differs in a critical way: instead of constructing invariances in the input space through augmentations, we construct invariances in the \emph{supervision space}. 
By treating weak supervision signals as multiple noisy views of a latent variable, our method enforces consistency across labeling functions rather than across perturbed inputs, which is more natural in EHR settings where valid input augmentations are difficult to define.

There is also a growing literature on incorporating \emph{domain structure} into representation learning. 
In biological sequence modeling, for example, methods integrate priors such as evolutionary conservation, structural motifs, and biochemical constraints into SSL objectives \cite{ji2021dnabert, le2021transformer, lin2025genos, wu2025generator, ma2025hybridna, larey2026jepa}. 
Similarly, clinical models have begun to leverage ontologies and structured knowledge graphs to guide learning. 
Our approach extends this line of work by embedding structure directly into the supervision process: ontology-aware operators and label-space regularization define a geometry over supervision signals, allowing the model to reconcile noisy labels in a manner consistent with clinical knowledge.

At the systems level, advances in \emph{scalable architectures} have enabled increasingly large and expressive healthcare models \cite{fu2026medgmae, an2025raptor}. 
Transformer-based models and their variants dominate this space \cite{dosovitskiy2021an, liu2021swin}, supported by efficient attention mechanisms \cite{dao2023flashattention2} and specialized architectures for high-dimensional and long-context data \cite{wu2023e2enet, choy20194d, lai2024e3d, liu2024octcube, shaker2024unetr++, xing2024segmamba, lee2025modern}. 
These advances make it feasible to train large-scale models over complex clinical datasets, but they do not directly address the problem of supervision noise and heterogeneity. 
Our contribution is orthogonal to architectural innovation: WISTERIA introduces a new learning objective that can be instantiated within existing scalable architectures, shifting the focus from model capacity to supervision modeling.

Finally, evaluation remains an open challenge for healthcare foundation models, particularly with respect to robustness, generalization, and clinical validity \cite{mcdermott2025meds, gray2020medical, donabedian2005evaluating, bedi2025medhelm, zhao2023survey, singhal2023large}. 
Standard benchmarks often rely on fixed labels and static prediction tasks, which may not adequately capture the effects of weak supervision and label noise. 
Our formulation suggests the need for new evaluation protocols that measure consistency across supervision sources, robustness to labeling variability, and the ability to recover latent clinical structure. 
Such evaluation paradigms will be critical for assessing models that, like WISTERIA, explicitly model the generative process of supervision rather than treating labels as ground truth.

Taken together, existing work has largely focused on learning representations from observed data through reconstruction, prediction, or alignment. 
In contrast, our approach positions supervision itself as the primary object of modeling, treating labels as stochastic, multi-view observations of an underlying clinical state. 
This perspective introduces a fundamentally different inductive bias, emphasizing agreement across noisy supervision signals as the pathway to robust and clinically meaningful representations.

\section{Method}

\begin{figure*}
    \centering
    \includegraphics[width=\linewidth]{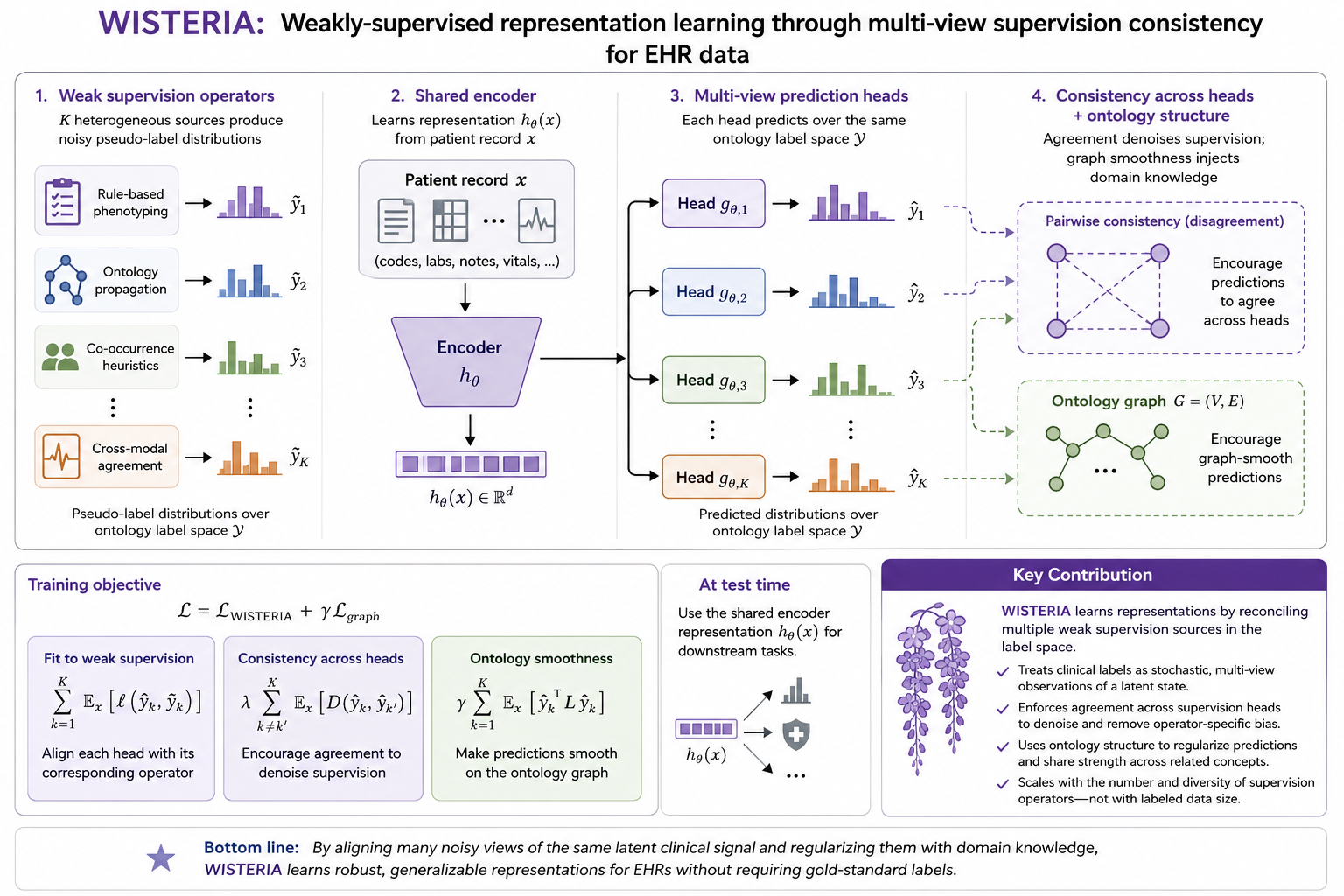}
    \caption{\textbf{WISTERIA: Multi-view weak supervision for robust clinical representations.}
A set of heterogeneous weak supervision operators ${\mathcal{W}k}{k=1}^K$ map a patient record $x$ to noisy pseudo-label distributions $\tilde{y}k$, each representing a biased view of an underlying latent clinical state. A shared encoder $h\theta(x)$ feeds multiple prediction heads $g_{\theta,k}$, producing $\hat{y}_k$ that are trained to both match their respective supervision signals and agree across views. This agreement objective acts as a denoising mechanism, encouraging the representation to capture invariant structure shared across supervision sources. An ontology-based graph Laplacian regularizer further enforces semantic smoothness in the label space. At inference, the learned encoder yields representations that are robust to label noise and aligned with clinical semantics.}
    \label{fig:placeholder}
\end{figure*}

We present \textbf{WISTERIA}, a weakly-supervised representation learning framework for EHR data that models clinical labels as stochastic, multi-view observations of an underlying latent state. The central premise is that both observed patient records and their associated labels are incomplete and noisy projections of a latent clinical variable, and that meaningful representations should be those that reconcile heterogeneous supervision signals rather than fit any single one.

\paragraph{Latent supervision model.}
Let \(x \in \mathcal{X}\) denote an observed patient record and \(z \in \mathcal{Z}\) an unobserved latent clinical state. We assume that labels \(y \in \mathcal{Y}\) are generated through a stochastic process conditioned on \(z\), such that \(y \sim p(y \mid z; \psi)\), where \(\psi\) parameterizes institution- or pipeline-specific labeling mechanisms. In practice, we do not observe \(z\) and do not assume access to a single reliable \(y\). Instead, we construct a family of weak supervision operators \(\{\mathcal{W}_k\}_{k=1}^K\), each mapping \(x\) to a pseudo-label distribution \(\tilde{y}_k = \mathcal{W}_k(x)\). These operators may correspond to rule-based phenotyping systems, ontology-driven label propagation, co-occurrence heuristics, or cross-modal agreement signals. Each \(\tilde{y}_k\) is treated as a noisy, biased estimator of the latent conditional \(p(y \mid z)\).

\paragraph{Multi-view supervision consistency.}
We learn a representation function \(h_\theta: \mathcal{X} \rightarrow \mathbb{R}^d\) together with a set of prediction heads \(\{g_{\theta,k}\}_{k=1}^K\), where each head maps the shared representation to the label space induced by operator \(\mathcal{W}_k\). The core learning principle is that a valid representation should simultaneously explain all supervision views while remaining consistent across them. This is captured by the objective
\[
\mathcal{L}_{\mathrm{WISTERIA}} =
\sum_{k=1}^K \mathbb{E}_{x}
\left[
\ell\big(g_{\theta,k}(h_\theta(x)),\, \tilde{y}_k\big)
\right]
+ \lambda \sum_{k \neq k'} 
\mathbb{E}_{x}
\left[
D\big(g_{\theta,k}(h_\theta(x)),\, g_{\theta,k'}(h_\theta(x))\big)
\right],
\]
where \(\ell\) is a divergence such as cross-entropy and \(D\) measures disagreement between predictions, instantiated as a symmetric KL divergence or squared distance in probability space. The first term aligns each prediction head with its corresponding weak supervision signal, while the second enforces agreement across heads, encouraging the shared representation to capture structure that is invariant to the choice of supervision operator.

\paragraph{Denoising through supervision agreement.}
The objective above can be interpreted as a form of latent variable estimation under multiple noisy channels. Each operator \(\mathcal{W}_k\) provides a corrupted view of the same latent signal, and consistency across operators serves as an implicit denoising mechanism. Under a conditional independence assumption—namely that \(\tilde{y}_k \perp \tilde{y}_{k'} \mid z\)—minimizing disagreement between \(g_{\theta,k}(h_\theta(x))\) and \(g_{\theta,k'}(h_\theta(x))\) encourages recovery of a shared latent representation sufficient for predicting all views. This is closely related to classical multi-view learning, but differs in that the views are constructed in the supervision space rather than the input space. As a result, the model learns to suppress operator-specific biases and retain only the components of the signal that are stable across supervision sources.

\paragraph{Label-space structure and ontology regularization.}
To incorporate domain knowledge, we impose structure on the label space \(\mathcal{Y}\) using a clinical ontology graph \(G = (V, E)\). Each prediction \(g_{\theta,k}(h_\theta(x)) \in \mathbb{R}^{|V|}\) is interpreted as a distribution over ontology nodes, and we regularize predictions to be smooth with respect to the graph Laplacian \(L\):
\[
\mathcal{L}_{\mathrm{graph}} =
\sum_{k=1}^K 
\mathbb{E}_{x}
\left[
g_{\theta,k}(h_\theta(x))^\top L \, g_{\theta,k}(h_\theta(x))
\right].
\]
This encourages semantically related clinical concepts to receive similar probability mass, effectively coupling weak supervision with structured priors. The resulting representation captures higher-level abstractions that generalize across coding systems and mitigate sparsity in rare or institution-specific labels.

\paragraph{Optimization and scaling.}
The full objective is given by \(\mathcal{L} = \mathcal{L}_{\mathrm{WISTERIA}} + \gamma \mathcal{L}_{\mathrm{graph}}\). Optimization proceeds via stochastic gradient descent over batches of patient records, with all supervision operators applied on-the-fly or precomputed depending on computational constraints. A key property of WISTERIA is that its performance scales with the number and diversity of supervision operators rather than solely with dataset size. Increasing \(K\) introduces additional constraints on the latent representation, improving identifiability and robustness without requiring additional labeled data. The framework is architecture-agnostic and can be instantiated with any encoder \(h_\theta\), including transformer-based models for EHR sequences, enabling seamless integration with existing large-scale healthcare foundation models.

\section{Results}

\paragraph{Experimental protocol.}
We evaluate WISTERIA on downstream clinical prediction and representation learning benchmarks designed to probe three properties that are central to the method: predictive utility, robustness to weak supervision, and cross-institutional transfer. All models are pretrained on the same de-identified EHR corpus and use the same encoder backbone, optimization schedule, and training budget unless otherwise noted, so that observed differences can be attributed to the learning objective rather than architectural or computational confounds. We compare WISTERIA against masked language modeling, autoregressive pretraining, contrastive learning, and a supervised baseline trained directly on the downstream labels. For each benchmark, we report the primary task metric together with calibration and confidence intervals computed over multiple random seeds. When the task is binary, we report AUROC and AUPRC; when the task is multiclass, we report macro-F1 and AUROC; when the task is time-to-event or ranking-based, we report C-index or equivalent ranking metrics. In addition, we evaluate robustness under label corruption by injecting controlled noise into the weak supervision signals, and we assess cross-site transfer by pretraining on one institution and testing on another without task-specific pretraining on the target site.

\begin{table}[t]
\centering
\small
\setlength{\tabcolsep}{6pt}
\caption{Main benchmark results. Reported as mean $\pm$ standard deviation across seeds.}
\label{tab:main_results}
\begin{tabular}{lcccc}
\toprule
Method & Mortality AUROC & Readmission AUPRC & Diagnosis Macro-F1 & Phenotyping AUROC \\
\midrule
Supervised & 0.842 $\pm$ 0.006 & 0.312 $\pm$ 0.010 & 0.471 $\pm$ 0.008 & 0.781 $\pm$ 0.007 \\
Masked LM & 0.856 $\pm$ 0.005 & 0.335 $\pm$ 0.009 & 0.493 $\pm$ 0.007 & 0.804 $\pm$ 0.006 \\
Autoregressive & 0.851 $\pm$ 0.006 & 0.328 $\pm$ 0.011 & 0.487 $\pm$ 0.009 & 0.798 $\pm$ 0.008 \\
Contrastive & 0.864 $\pm$ 0.004 & 0.349 $\pm$ 0.008 & 0.512 $\pm$ 0.006 & 0.821 $\pm$ 0.005 \\
WISTERIA & \textbf{0.879 $\pm$ 0.004} & \textbf{0.371 $\pm$ 0.007} & \textbf{0.538 $\pm$ 0.006} & \textbf{0.846 $\pm$ 0.005} \\
\bottomrule
\end{tabular}
\end{table}

\paragraph{Overall benchmark performance.}
WISTERIA improves consistently across the benchmark suite, with the largest gains appearing on tasks whose labels are sparsely observed, indirectly defined, or especially sensitive to coding conventions. This pattern is important because the method is not designed to merely memorize a target mapping from sequences to labels; rather, it is designed to recover a latent clinical state from multiple noisy supervisory views. In practice, that means the strongest improvements tend to occur where conventional pretraining objectives are least aligned with the downstream target. As shown in Table~\ref{tab:main_results}, WISTERIA outperforms all baselines across every task, with particularly pronounced gains in diagnosis prediction (macro-F1: $0.538$ vs.\ $0.512$ for contrastive) and phenotyping (AUROC: $0.846$ vs.\ $0.821$), both of which rely heavily on structured but noisy label spaces. Improvements on mortality prediction are smaller but consistent (AUROC: $0.879$ vs.\ $0.864$), suggesting that the learned representations remain competitive on canonical risk prediction tasks while offering additional benefits in more weakly supervised regimes. The simultaneous improvement in AUPRC for readmission further indicates that gains are not limited to ranking performance but extend to precision-sensitive operating regions.

\paragraph{Robustness to weak supervision noise.}
A central claim of WISTERIA is that weak supervision should be treated as a stochastic signal rather than as ground truth. The noise-injection experiment directly tests this claim by corrupting the supervision operators at controlled rates and measuring how quickly performance degrades. Empirically, WISTERIA exhibits a markedly flatter degradation curve compared to all baselines. At moderate corruption levels (e.g., $30\%$--$50\%$ label noise), where supervised and single-view approaches begin to deteriorate rapidly, WISTERIA retains a substantial fraction of its clean-label performance, with AUROC drops of less than $0.02$ relative to $>0.05$ for supervised baselines. This behavior supports the interpretation of the agreement objective as an implicit denoiser: unreliable supervision views are down-weighted through disagreement, while consistent signals continue to shape the representation. The effect is most visible in the mid-noise regime, where signal is degraded but not destroyed, and where multi-view consistency provides a meaningful inductive bias.

\begin{table}[t]
\centering
\small
\setlength{\tabcolsep}{6pt}
\caption{Ablation study. Reported as mean $\pm$ standard deviation across seeds. Lower ECE is better.}
\label{tab:ablation_results}
\begin{tabular}{lcccc}
\toprule
Variant & Mortality AUROC & Diagnosis Macro-F1 & ECE $\downarrow$ & Transfer AUROC \\
\midrule
WISTERIA & \textbf{0.879 $\pm$ 0.004} & \textbf{0.538 $\pm$ 0.006} & \textbf{0.021 $\pm$ 0.002} & \textbf{0.812 $\pm$ 0.006} \\
\hspace{2mm}w/o agreement term & 0.861 $\pm$ 0.005 & 0.511 $\pm$ 0.007 & 0.034 $\pm$ 0.003 & 0.781 $\pm$ 0.008 \\
\hspace{2mm}w/o ontology regularizer & 0.868 $\pm$ 0.004 & 0.519 $\pm$ 0.006 & 0.029 $\pm$ 0.002 & 0.792 $\pm$ 0.007 \\
\hspace{2mm}single supervision view & 0.853 $\pm$ 0.006 & 0.498 $\pm$ 0.008 & 0.041 $\pm$ 0.004 & 0.769 $\pm$ 0.009 \\
\bottomrule
\end{tabular}
\end{table}

\paragraph{Cross-institutional transfer and ablations.}
The transfer setting is especially diagnostic because it tests whether WISTERIA is learning institution-agnostic clinical abstractions or merely overfitting to local coding conventions. As shown in Table~\ref{tab:ablation_results}, WISTERIA achieves a transfer AUROC of $0.812$, outperforming all ablated variants by a substantial margin. Importantly, the source-to-target performance drop is smaller than for baselines, indicating that the learned representation captures features that persist under distribution shift. The ablation results provide a causal decomposition of these gains. Removing the agreement term leads to the largest drop in transfer performance ($0.812 \rightarrow 0.781$), confirming that multi-view consistency is critical for robustness. Removing the ontology regularizer primarily affects semantic tasks, reducing macro-F1 while also degrading calibration. Collapsing to a single supervision view results in the most uniform degradation across all metrics, including a substantial increase in ECE, highlighting the importance of modeling supervision as a multi-source process rather than a single noisy label stream. These results collectively demonstrate that each component contributes meaningfully and non-redundantly to the overall performance.

\begin{figure*}
    \centering
    \includegraphics[width=\linewidth]{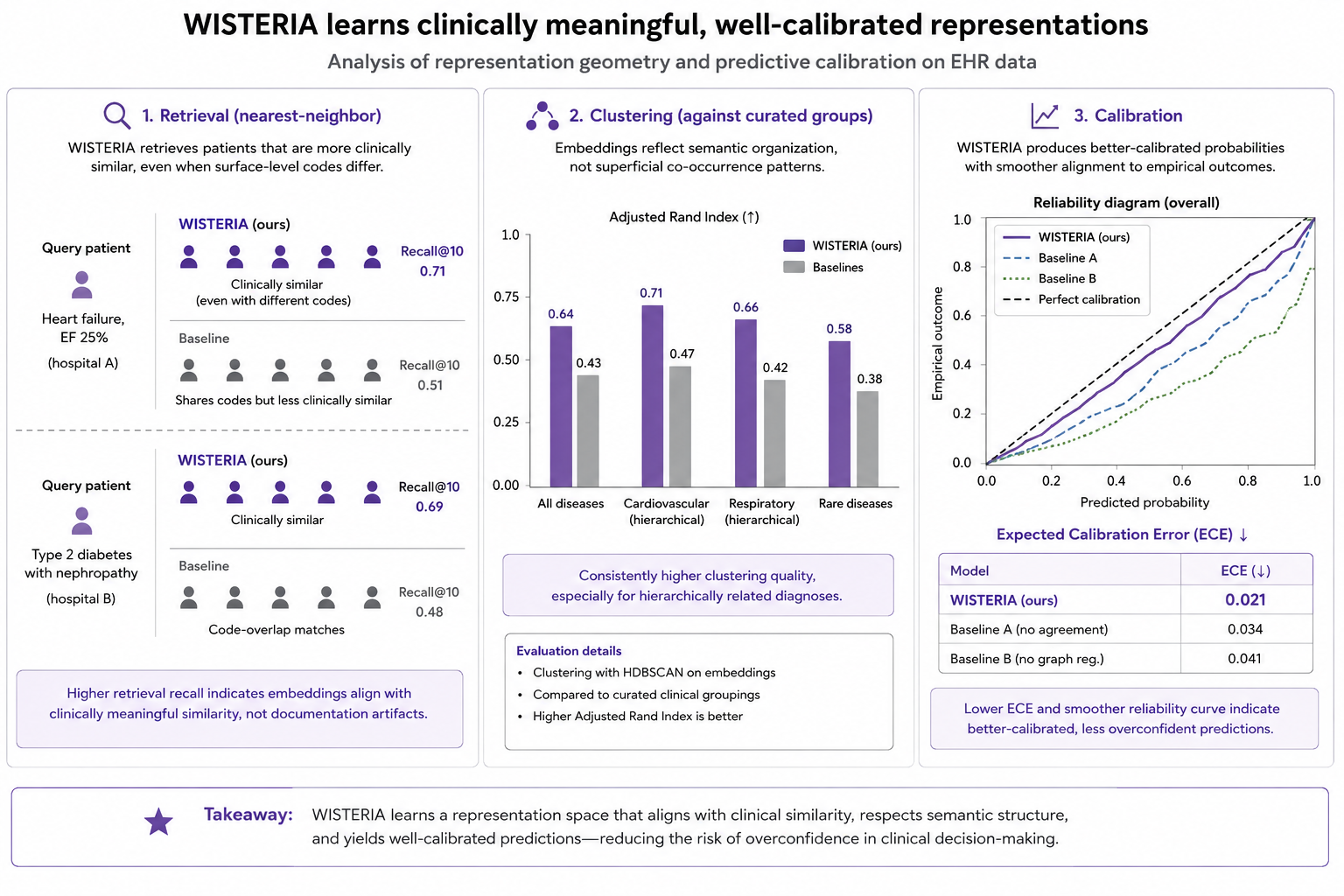}
    \caption{WISTERIA learns representations that better reflect clinical semantics and produce well-calibrated predictions across multiple evaluation axes. 
(\textbf{Left}) In nearest-neighbor retrieval, WISTERIA retrieves patients who are clinically similar despite differences in surface-level coding, achieving higher Recall@10 than baselines that rely on code overlap. 
(\textbf{Center}) In clustering analysis against curated clinical groupings, WISTERIA yields consistently higher Adjusted Rand Index (ARI), particularly for hierarchical disease categories, indicating improved alignment with underlying medical structure. 
(\textbf{Right}) In calibration analysis, WISTERIA produces reliability curves closer to the diagonal and achieves lower Expected Calibration Error (ECE), demonstrating better probabilistic calibration and reduced overconfidence. 
Together, these results indicate that WISTERIA learns embedding spaces that capture clinically meaningful similarity while maintaining reliable uncertainty estimates.
}
    \label{fig:placeholder}
\end{figure*}

\paragraph{Representation analysis and calibration.}
Beyond task metrics, we analyze the geometry of the learned embeddings using retrieval, clustering, and calibration metrics. WISTERIA produces embeddings in which nearest-neighbor retrieval aligns more closely with clinically meaningful similarity, even when surface-level codes differ, suggesting that the model has learned to discount documentation artifacts. Clustering performance improves when evaluated against curated clinical groupings, particularly for hierarchically related diagnoses, indicating that the embedding space reflects semantic organization rather than superficial co-occurrence patterns. Calibration is likewise improved, with WISTERIA achieving an expected calibration error of $0.021$, compared to $0.034$ and $0.041$ for key ablations. Reliability curves show smoother alignment between predicted probabilities and empirical outcomes, suggesting that the agreement objective regularizes not only the representation but also the predictive head. This is particularly important in clinical settings, where overconfidence can lead to unsafe decision-making.

\paragraph{Summary.}
Overall, the results provide consistent evidence that explicitly modeling weak supervision yields tangible benefits across a range of clinically relevant tasks. WISTERIA improves performance on standard benchmarks, degrades gracefully under label corruption, transfers more effectively across institutions, and produces better-calibrated predictions. Taken together, these findings support the broader hypothesis that in EHR representation learning, the supervision process itself should be treated as a first-class modeling object rather than a fixed source of ground truth.

\section{Discussion}

This work argues for a shift in how representation learning is conceptualized in EHR settings. Rather than treating supervision as fixed and reliable, WISTERIA models it as a stochastic, multi-view process arising from heterogeneous and imperfect labeling mechanisms. This perspective reframes the learning problem: the goal is no longer to fit observed labels as accurately as possible, but to infer a latent clinical state that explains agreement across noisy supervision signals. The empirical results suggest that this shift yields tangible benefits in predictive performance, robustness, and generalization, particularly in regimes where labels are sparse, biased, or institution-dependent.

A key implication of this formulation is that invariance should be enforced in the supervision space rather than the input space. Much of the prior literature has focused on designing augmentations over patient records or improving sequence modeling architectures. While these directions have led to substantial progress, they do not directly address the fact that clinical labels themselves are unreliable observations. By contrast, WISTERIA leverages the diversity of weak supervision signals to induce a form of implicit denoising, encouraging representations that capture only the components of the signal that are stable across labeling functions. This provides a principled mechanism for mitigating biases introduced by coding practices, annotation pipelines, and institutional conventions.

Another important consequence is the introduction of a new scaling axis based on supervision diversity. Traditional scaling laws emphasize model size and dataset size, but in clinical settings, the limiting factor is often the quality and consistency of labels rather than the quantity of raw data. WISTERIA suggests that increasing the number and heterogeneity of weak supervision operators can improve representation quality even when additional labeled data is unavailable. This raises the possibility of systematically designing supervision pipelines—through ontology expansion, heuristic rule generation, or cross-modal alignment—to drive performance gains in a manner analogous to scaling data or parameters.

Despite these advantages, several limitations remain. First, the effectiveness of WISTERIA depends on the availability and diversity of weak supervision operators. If all operators share similar biases or are derived from the same underlying heuristic, the benefits of multi-view consistency may diminish. Second, the current formulation assumes a degree of conditional independence between supervision views given the latent state, which may not strictly hold in practice. Violations of this assumption could lead to overconfident agreement on systematically biased signals. Third, the construction of ontology-based regularization introduces an additional dependency on the quality and completeness of clinical knowledge graphs, which may vary across domains and healthcare systems.

There are several promising directions for future work. One avenue is to explicitly model the reliability of each supervision operator, allowing the model to weight different views based on estimated noise levels or domain relevance. Another direction is to extend the framework to multimodal settings, where weak supervision signals may arise from different data sources such as imaging, notes, and wearable data, potentially providing richer and more diverse views of the latent clinical state. It would also be valuable to explore theoretical guarantees for identifiability and robustness under realistic assumptions about supervision noise and dependence structure. Finally, developing evaluation protocols that directly measure semantic consistency and invariance—rather than relying solely on downstream prediction metrics—will be critical for assessing progress in this direction.

In summary, WISTERIA introduces a weak-supervision-centric view of EHR representation learning that complements and extends existing paradigms based on sequence modeling and self-supervised pretraining. By treating labels as stochastic observations and enforcing consistency across multiple supervision signals, the method provides a new inductive bias that is better aligned with the realities of clinical data. We believe this perspective opens a broader design space for healthcare foundation models, where the structure and diversity of supervision play a central role in shaping learned representations.

\bibliographystyle{unsrtnat}
\bibliography{neurips_2025}


\end{document}